\documentclass[11pt,a4paper]{article}
\usepackage{times,latexsym}
\usepackage{amsmath}
\usepackage{url}
\usepackage[T1]{fontenc}
\usepackage{adjustbox}
\usepackage{graphicx}
\usepackage{comment}
\usepackage{multirow}
\usepackage{booktabs}
\usepackage{float}
\usepackage{fancyhdr}
\usepackage{ifthen}
\graphicspath{ {figures/} }

\usepackage[acceptedWithA]{tacl2018v2}
\usepackage{xspace,mfirstuc,tabulary}

\usepackage{dsfont}

\newif\iftaclinstructions
\taclinstructionsfalse % AUTHORS: do NOT set this to true
\iftaclinstructions
\renewcommand{\confidential}{}
\renewcommand{\anonsubtext}{(No author info supplied here, for consistency with
TACL-submission anonymization requirements)}
\newcommand{\instr}
\fi

\iftaclpubformat % this "if" is set by the choice of options

\else

\fi

\title{On Generative Spoken Language Modeling from Raw Audio}

\author{Kushal Lakhotia\Thanks{equal contribution. $^\ddagger$ Also at EHESS. $^\dagger$ Also at INRIA.  $^\mathsection$ Work done while at FAIR.} , Eugene Kharitonov$^*$, Wei-Ning Hsu, Yossi Adi, Adam Polyak, \\ \textbf{Benjamin Bolte$^{\mathsection}$, Tu-Anh Nguyen$^{\dagger}$, Jade Copet, Alexei Baevski,} \\ \textbf{Abdelrahman Mohamed, Emmanuel Dupoux$^{\ddagger}$}\\
  Facebook AI Research      \\  \\ {\tiny Author's version of   \tiny Lakhotia, K., Kharitonov, E., Hsu, W.-N., Adi, Y., Polyak, A., Bolte, B., Nguyen, T.-A., Copet, J., Baevski, A., Mohamed, A., \& Dupoux, E. (2022). \vspace{-1em}} \\  {\tiny On Generative Spoken Language Modeling from Raw Audio, \textit{Transactions of the Association for Computational Linguistics}}
  }

\date{}

\newcommand{\urldemo}[0]{https://speechbot.github.io/gslm}
\newcommand{\urlrepo}[0]{https://github.com/pytorch/fairseq/tree/master/examples/textless_nlp/gslm}

\begin{document}
\thispagestyle{fancy}
\maketitle

%\pagestyle{fancy}
%\fancyhf{}
%\rhead{\tiny Overleaf}

\begin{abstract}
We introduce \textit{Generative Spoken Language Modeling}, the task of learning the acoustic and linguistic characteristics of a language from raw audio (no text, no labels), and a set of metrics to automatically evaluate the learned representations at acoustic and linguistic levels for both encoding and generation. %For the generative mode, we introduce two end-to-end tasks: speech resynthesis (repeating the speech input using the system's own voice), and  speech generation (producing novel speech outputs conditional on a spoken prompt, or unconditionally). 
We set up baseline systems consisting of a discrete speech encoder (returning pseudo-text units), a generative language model (trained on pseudo-text), and a speech decoder (generating a waveform from pseudo-text) all trained without supervision and validate the proposed metrics with human evaluation. Across 3 speech encoders (CPC, wav2vec 2.0, HuBERT), we find that the number of discrete units (50, 100, or 200) matters in a task-dependent and encoder-dependent way, and that some combinations approach text-based systems.\footnote{Evaluation code and trained models are here:  {\scriptsize{\url{\urlrepo}}}. \\ Sample audios are here: \scriptsize\url{\urldemo}.}

\end{abstract}

\begin{table*}[ht]
    \centering
    \small
    \begin{tabular}{lp{1.5cm}p{2.5cm}cp{1.5cm}p{4cm}p{1.5cm}}
    \hline
        & \multicolumn{2}{c}{Encoding} && \multicolumn{3}{c}{Generation} \\
        \cline{2-3}
        \cline{5-7}
        Level& Task&\multicolumn{1}{l}{Automatic metric} && Task&\multicolumn{1}{c}{Automatic metric}& Human \\
        \midrule
        Language 
        & Spoken LM
        & \textbf{Spot-the-word}, Syntax-Acc
           && Speech Gen. &\textbf{AUC-of-VERT/PPX}, cont-BLEU, PPX@o-VERT
           & \bf MMOS \\
                   \midrule

        Acoustic
           & Acoustic Unit Disc.
           & \textbf{ABX-across}, ABX-within        
           && Resynthesis&\textbf{PER-from-ASR}, CER-from-ASR
           & CER, \bf MOS \\
           \hline
    \end{tabular}
    \caption{\textbf{Tasks and metrics} proposed to evaluate encoding/generation quality of models at the acoustic or language levels. Bold fonts highlights the main metric used for each category (Section~\ref{sec:methods} for details).}
    \label{tab:metric_overview}
    \vspace{-0.8em}
\end{table*}

\vspace{-0.2cm}
\section{Introduction}
\vspace{-0.2cm}

An open question for AI research is creating systems that learns from natural interactions as infants learn their first language(s): from raw uncurated data, and without access to text or expert labels \cite{dupoux2018cognitive}. Natural Language Processing (NLP) systems are currently far from this requirement. Even though great progress has been made in reducing or eliminating the need for expert labels through self-supervised training objectives ~\cite{brown2020language, peters-etal-2018-deep,radford2019language,  devlin-etal-2019-bert, liu2019roberta, NEURIPS2019_c20bb2d9, lewis-etal-2020-bart}, the basic units on which these systems are trained are still textual. Yet, young children learn to speak several years before they can read and write, providing a proof of principle that language can be learned without any text. Being able to achieve 'textless NLP' would be beneficial for the majority of the world's languages which do not have large textual resources or even a widely used standardized orthography (Swiss German, dialectal Arabic, Igbo, etc.), and which, despite being used by millions of users, have little chance of being served by current text-based technology. It would also be useful for 'high-resource' languages, where the oral and written forms often mismatch in terms of lexicon and syntax, and where some linguistically relevant signals carried by prosody and intonation are basically absent from text. While text is still the dominant form of language on the web, a growing amount of audio resources like podcasts, local radios, social audio apps, on-line video games provide the necessary input data to push NLP to an audio-based future and thereby expand the inclusiveness and expressivity of AI systems.

% Some early investigations have shown possible to learn representations that capture 

Is it possible to build an entire dialogue system from audio inputs only? This is a difficult challenge, but breakthroughs in unsupervised representation learning may address part of it. Unsupervised learning techniques applied to speech were shown to learn continuous or discrete representations that capture speaker invariant phonetic content \cite{versteegh2016zero,dunbar2020zero}, despite themselves not being phonemic \cite{schatz2021}. Recent developments in self-supervised learning have shown impressive results as a pretraining technique \cite{van2017neural,chung2019unsupervised,hsu2020hubert}, to the extent that Automatic Speech Recognition (ASR) on par with the state of the art from two years back can be built with 5000 times less labelled speech \cite{baevski2020wav2vec}, or even no with no labelled speech at all \cite{baevski21}. Of course, ASR still assumes access to text to learn a language model (LM) and the mapping  to the audio units. Here, we study the case where the LM is directly trained from the audio units without any recourse to text. 
%Setting textless NLP as a north star would enable us to develop more inclusive AI, and to enrich existing NLP applications with the full expressivity of spontaneous oral language. This is also timely given the fast growing presence of non annotated audio content on the web (podcasts, local radio, social audio apps and video games). 
%Yet, these techniques only remain available to languages with large textual resources, leaving out thousands of languages which, while used by millions of people, do not have a widely used written form (e.g., Swiss German, Igbo, many dialects of Arabic, etc.). Being able to train Language Models (LMs) from raw audio only would go a long way towards making language technologies more inclusive. It would also be useful in high resourced languages, where the written and oral forms mismatch at many linguistic levels. %Training generative LMs from raw audio would open up the possibility of downstream applications like speech to speech translation or dialogue systems that would be more expressive and closer to spoken language.  
%Fortunately,
%Recent advances in unsupervised representation learning from audio \cite{chung2019unsupervised,baevski2020wav2vec,kharitonov2020data,hsu2020hubert} make it possible to envision generative spoken language modeling. 

\begin{figure}
    \centering
    \includegraphics[width=0.5\textwidth]{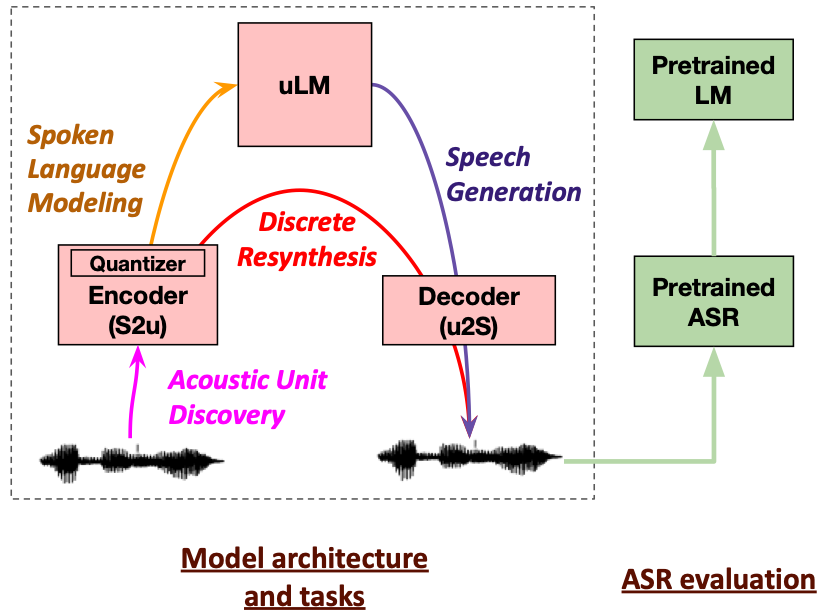}
    \caption{\textbf{Setup of the baseline model architecture, tasks and metrics.}  %A discrete encoder converts Speech into discrete pseudo-text Units (S2U), the Unit-based Language Model (ULM) models the distribution of the pseudo-text, the decoder converts Units back to waveform (U2S). Acoustic Unit Discovery learns the pseudo-text units from speech, Discrete Resynthesis converts speech into speech via a pseudo-text bottleneck, Spoken Language Modeling learns the probability distribution of the input speech, Speech Generation samples new speech utterances conditionally or unconditionally. The automatic tests for the generative tasks use a pretrained ASR to convert the wav back into text and a pretrained LM to compute text-domain metrics.
    %The metrics: they measure either the quality of the pseudo-text acoustic units (bottom) or of the language model built on top of it (top), in either encoding tasks (left) or generative tasks (right), and are either automatic (black) or based on human judgments (blue). 
    }
    \label{fig:setup}
    \vspace{-0.7em}
\end{figure}

The high level idea (see Figure~\ref{fig:setup}) is that automatically discovered discrete units can be used to encode speech into "pseudo-text" (speech-to-unit, S2u), which is used in turn to train a generative language model (unit-based language model, uLM) and to train a speech synthesizer (unit-to-speech, u2S). This enables learning an LM from scratch without text, and use it to generate speech conditionally or unconditionally, essentially replicating what toddlers achieve before learning to read. Early studies using discrete codes learned from an autoencoder
show the feasibility of such an approach, but remain at a level of a demo \cite{van2017neural}.%\footnote{\href{https://avdnoord.github.io/homepage/vqvae/}{avdnoord.github.io/homepage/vqvae/}} \ap{Consider removing this sentence + remove link, we credit vqvae in related work already. Alternatively, remove the url... }.

In this paper, we address one major conceptual stumbling block which has, thus far, prevented such early studies from having the transformative impact they could have in language technology: \textit{model evaluation}. We contend that it will be impossible to make progress in this area beyond demos unless proper evaluation methods enabling system comparison are etablished. 

Evaluation for speech generation is difficult due to the continuous, variable and multi-level nature of the speech waveform, and the necessity both to capture fine grained acoustic details to generate  intelligible audio and to abstract away from them to learn higher level language concepts. Text-based models do not have this problem, since the input is already expressed in terms of mid-level discrete units (characters or words), and are typically evaluated with unsupervised metrics close to the learning objectives like perplexity or log likelihood. Here, such an approach is not directly applicable even if we rely on discrete pseudo-text units, since such metrics would depend in an unknown fashion on their granularity (number, duration and distribution), making the comparison of models that use different units infeasible. %In fact, some speech generation models may not have any discrete units at all, making model-independent perplexity measures even more difficult to define.

 Conceptually, generative spoken language models can be evaluated at two levels, the acoustic and the language levels, and through two modes of operation, encoding and generation, resulting in 2x2 tasks (see  Table \& Figure~\ref{tab:metric_overview}). \textit{Acoustic Unit Discovery} (encoding at the acoustic level) consists in representing speech in terms of discrete units discarding non-linguistic factors like speaker and noise. \textit{Spoken Language Modeling} (encoding at the language level) consists in learning the probabilities of language patterns. \textit{Speech Resynthesis} (generation for acoustic modeling) consists in generating audio from given acoustic units.  This boils down to repeating in a voice of choice an input linguistic content encoded with speech units. \textit{Speech Generation} (generation for language modeling) consists in generating novel and natural speech (conditioned on some prompt or not). Compared to standard text generation, a critical and novel component of the audio variant is clearly the discovery of units since it conditions all the other components. This is why we devote our analyses of model architectures to the unit-to-speech component specifically, and leave it for further work to evaluate how the downstream components can also be optimized for spoken language generation.

The major contributions of this paper are as follows : (1) we introduce two novel evaluation metrics for the generation mode of spoken language modeling at the acoustic and language levels respectively. Our key insight is to use a generic pretrained ASR system to establish model-independent assessments of the intelligibility (acoustic level) and meaningfulness (language level) of the produced outputs. The ASR system converts the generated waveform back to text, enabling us to adapt standard text-based metrics for these two levels. (2) we validate these metrics through comparison with human evaluation. We show a high degree of concordance between human and machine evaluations of intelligibility and meaningfulness of generated audio. (3) we show that these metrics can be predicted by simpler ones geared to evaluate the encoding mode of the spoken LM. Zero-shot metrics borrowed from previous studies in the Zero Resource Speech Challenges \cite{versteegh2016zero,Nguyen2020} correlate well with their generative counterpart, offering an easier proxy to rapidly iterate on model selection. (4) we systematically study the effect of the type of encoding units by factorially crossing three recent speech-to-unit encoders, CPC, Wave2vec 2.0 and HuBERT, with three codebook sizes for the discrete units, 50, 100, 200. We keep constant the rest of the system built from out-of-the-box components (standard Transformer for the uLM, Tacotron 2 for u2S). We show that both the encoder type and the number of units matter, and that they matter differently depending on the evaluation task. (5) we open source our evaluation tools and models to help reproducibility and comparability with future work.

In Section \ref{sec:methods}, we introduce the ASR, zero-shot and human evaluation metrics, in Section \ref{sec:models} we present the models, in  
Section \ref{sec:results}, we analyze the results and discuss them in Section \ref{sec:discuss}.

%To illustrate our metrics, we construct baseline pipeline models made of three components (Figure \ref{fig:setup}): a speech-to-unit (S2U) Encoder, which takes as input a waveform and outputs discrete units (Section~\ref{sec:encoders}), a unit Language Model (uLM), which learns the probability distribution of the learned unit sequences (Section \ref{sec:LMs}), and a unit-to-speech (U2S) Decoder which generates a waveform from a unit sequence (Section \ref{sec:pTTS}). We use state-of-the-art systems for each of these components. Such a pipeline can be setup easily for our two generative tasks: speech resynthesis (S2U $\rightarrow$ U2S), and speech generation (ULM $\rightarrow$ U2S). The latter is tested unconditionally or conditioned on a spoken prompt which seeds the ULM to continue the fragment or generate a follow-up sentence. This can be viewed as an embryonic dialogue system. We compare these baselines to equivalent ``toplines'' trained with text where automatic speech recognition (ASR), character LM, and text-to-speech (TTS) models are S2U, ULM, and U2S models respectively. %This provides a ``topline'', trained on the same data, against which to compare our speech only systems.
%In Section \ref{sec:results}, we compare 3 different encoders (\namemodels{}), from which we extract codebooks of different sizes (50, 100, and 200 units) while keeping the LM and decoder constant, and discuss the results in Section \ref{sec:discuss}.

\vspace{-.4em}

\section{Related work}\label{sec:related}
\vspace{-.2em}

\noindent{\textbf{Unsupervised speech representation learning}}  aims to distill features useful for downstream tasks, such as phone discrimination~\cite{kharitonov2020data,schneider2019wav2vec} and semantic prediction~\cite{lai2021semi,wu2020self}, by constructing pretext tasks that can exploit large quantities of unlabeled speech. Pretext tasks in the literature can be roughly divided into two categories: reconstruction and prediction. Reconstruction is often implemented in the form of auto-encoding~\cite{hsu2017learning}, where speech is first encoded into a low-dimensional space, and then decoded back to speech. Various constraints can be imposed on the encoded space, such as temporal smoothness~\cite{ebbers2017hidden,glarner2018full,khurana2019factorial,khurana2020convolutional}, discreteness~\cite{ondel2016variational,van2017neural}, and presence of hierarchy~\cite{lee2012nonparametric,hsu2017unsupervised}.
Prediction-based approaches task a model with predicting information of unseen speech based on its context. Examples of information include spectrograms~\cite{chung2019unsupervised,wang2020unsupervised,chi2020audio,liu2020mockingjay,chung2020improved,liu2020non,ling2020deep,ling2020decoar}, cluster indices~\cite{baevski2019effectiveness,hsu2020hubert}, derived signal processing features~\cite{pascual2019learning,ravanelli2020multi}, and binary labels of whether a candidate is the target unseen spectrogram~\cite{oord2018representation,schneider2019wav2vec,baevski2019vq,kharitonov2020data,baevski2020wav2vec}.

\noindent{\textbf{Speech resynthesis.}} Recent advancements in neural vocoders enabled generating natural sounding speech and music~\cite{oord2016wavenet, melgan, kong2020hifi}. These are often conditioned on the log mel-spectrogram for the generation process. Learning low bitrate speech representations in an unsupervised manner, has attracted attention from both the machine learning and the speech communities~\cite{liu2019unsupervised, feng2019combining, nayak2019virtual, tjandra2019vqvae, schneider2019wav2vec, baevski2019vq, chen2020unsupervised, morita2020exploring, tobing2020cyclic}. These representations can later be used for generation without text, which is particularly important for low-resource languages~\cite{dunbar2019zero, dunbar2020zero}. \citet{van2017neural} proposed a Vector-Quantized Variational Auto-Encoder (VQ-VAE) model to learn discrete speech units, which will be later used for speech synthesis using a WaveNet model. \citet{eloff2019unsupervised} suggested a VQ-VAE model followed by a FFTNet vocoder model~\cite{jin2018fftnet}. \citet{tjandra2020transformer} suggested to use transformer~\cite{trans} together with a VQ-VAE model for unsupervised unit discovery, and \citet{van2020vector} combines vector quantization together with contrastive predictive coding for acoustic unit discovery. Another line of work use representations from an ASR acoustic model that are combined with identity and prosodic information for voice conversion~\cite{polyak2019tts, polyak2020unsupervised, polyak2021high}. In terms of evaluation, the Zero-Resource challenge~\cite{dunbar2019zero, dunbar2020zero, Nguyen2020} used bitrate together with human evaluation. In this paper we additionally introduce an ASR based evaluation metric.

%\paragraph{Text generation.} Text generation uses autoregressively trained LMs (lots of references) and are typically evaluated using measures of text quality (perplexity) plus some accompanying measure of diversity. Here, we adapt these techniques using an ASR to turn the generated audio into text. 

\vspace{-.5em}

\section{Evaluation Methods}\label{sec:methods}
\vspace{-.2em}

% emphasize the first set is novel and applicable to and GSLM systems
We present two sets of automatic evaluation metrics; the first ones assess the output of generative speech models (ASR metrics, Section \ref{sec:eeeval}); the second ones, the encoded representations (zero-shot probe metrics, Section \ref{sec:bbeval}). Finally, we present the human evaluations (Section \ref{sec:heval}).

%\paragraph{Syntactic level: acceptability judgement.} Similar to the preceding task, acceptability judgments consist in spotting the ungrammatical sentence out of pairs like ``Cats chase mice'' and ``Cats chases mice''. Here again this is done using (pseudo-) probabilities that can easily be extracted from LMs. Here we use the sBLIMP dataset~\cite{Nguyen2020} which is a speech rendering of the BLIMP dataset~\cite{Warstadt2020}.  As the preceding one, it has been turned into speech via Google API and filtered for phonological or prosodic counfounds, and restricted to the vocabulary present in the LibriSpeech 960 training set. 

% \vspace{-0.7em}

\subsection{Generation: ASR metrics}\label{sec:eeeval}
We present our new evaluation metrics for generation tasks. The first task, 
speech resynthesis, involves S2u which encodes input speech into units and u2S which decodes it back to speech. In this task, we wish to evaluate \textit{intelligibility} of the resulting speech. The second task, speech generation, involves the full S2u$\rightarrow$uLM$\rightarrow$u2S pipeline, and we wish to evaluate \textit{meaningfulness} of the generated speech. Our overall idea is to use ASR to convert the generated speech back to text and then use text-based metrics.

\noindent{\textbf{Speech resynthesis intelligibility: ASR-PER.}} The ideal metric for intelligibility would be to use humans to transcribe the resynthesized speech and compare the text to the original input. An automatic proxy can be obtained by using a state-of-the-art ASR system pretrained on a large corpus of real speech.\footnote{We use a \textsc{base} wav2vec 2.0 phoneme detection model trained on LibriSpeech-960h with CTC loss from scratch.} Our main metric is Phone Error Rate (PER), which only uses an acoustic-model ASR, without fusing with an additional language model \citep{chorowski2016towards}. In preliminary experiments we also experimented with a full ASR with an LM and computed Word Error Rate (WER) and Character Error Rate (CER) to give partial credit. The latter is probably closer to humans intelligibility metrics, as humans cannot turn off their lexicon or language model. We also computed such metrics by training a fitted ASR model for each resynthesis model on a specific training corpus (see Supplementary Section \ref{sup:fit}). The logic of this last test is that it provides a more direct measure of the information lost in the S2u$\rightarrow$u2S pipeline, because it could adapt to systematic errors introduced by the u2S model. Since the scores between these different approaches correlated highly, we only report here the results on the PER for a pretrained ASR model which is the simplest to deploy.

\noindent{\textbf{Speech generation quality and diversity: AUC on Perplexity and VERT.}} Text generation evaluation typically involves two axes: the quality of the generated text (with automatic metrics like mean perplexity or negative log likelihood computed on a reference large language model) and the diversity (with metrics like self-BLEU\footnote{Higher self-BLEU scores indicate lower diversity of the produced text.}, \citeauthor{zhu2018texygen}, \citeyear{zhu2018texygen}). Typically, there is a trade-off between these two dimensions based on the temperature hyperparameter used for sampling from the language model, whereby at low temperature, the system outputs good sentences but not varied, and at high temperatures, it outputs varied sentences, but not very good. This results in model comparison being either based on 2D plots with lines representing the trade-off between quality and diversity, or on aggregate metrics like the area under the curve. Preliminary explorations (see Appendix Section~\ref{sup:temp}) with our models revealed two problems preventing a straightforward application of such a scoring strategy. 

First, we found that for some models, at a low enough temperature, self-BLEU score stopped increasing, but the systems started to repeat more and more words within a sentence (e.g., ``the property the property the property''). We therefore introduce a new metric, auto-BLEU, that measures within-sentence diversity. For a single utterance $u$, auto-BLEU is calculated as the ratio of k-grams $s \in NG_k(u)$ that are repeated at least once:
\begin{equation}
\small
\texttt{auto-BLEU}(u, k) %mean_{n=1..N}(autobleu_n(u))\\
%\end{equation}
%\begin{equation}
%autobleu_n(u) 
= \frac{\sum_s \mathds{1} \left[s \in (NG_k(u) \backslash s) \right]}{| NG_k(n)|}
\end{equation}
As with BLEU score, to get n-gram auto-BLEU we calculate the geometric mean of $\texttt{auto-BLEU}(u, k)$ obtained for $k \in [1, n]$ and average over the set of generated utterances.
By calculating the geometric mean of self- and auto-BLEU, we obtain an aggregate metric which we call VERT (for diVERsiTy). We used a bigram version of self- and auto-BLEU.% (see Supplementary Section \ref{sup:VERT} for details).

Second, we found that critical temperatures for which the output was reasonable were not constant across models. This makes sense, because temperature controls the probability of sampling individual units, and the probabilistic distribution and duration of these units depend on the models. Here, we chose to use the oracle text as an anchor to compute reference temperatures, i.e., the temperatures at which the perplexity or the VERT score reach the values of the oracle text. 

\begin{figure}
    \centering
    \includegraphics[width=0.5\textwidth]{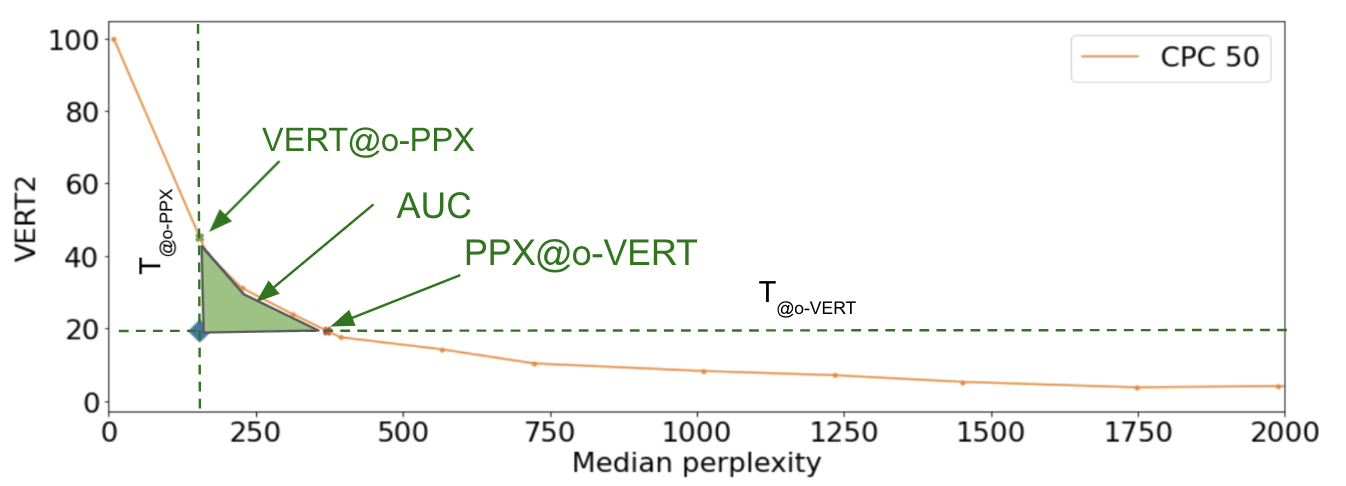}
    \caption{\textbf{Comparison of diversity and perplexity of the generated speech.}  We plot VERT vs. Median perplexity. The blue diamond corresponds to the oracle reference point. It defines two cut-offs on the curve: VERT @oracle-PPX and PPX @oracle-VERT. The green area corresponds to the AUC metric.} 
    % For reference, the yellow and red curves are for character LM and CPC-50 LM.
    \label{fig:AUC}
    \vspace{-.7em}
\end{figure}

This gives us boundary conditions at which we can compare (the perplexity at oracle diversity and the diversity at oracle perplexity), as well as a method to compute the area under curve (AUC) between these two boundaries (See Figure \ref{fig:AUC}). As AUC decreases, the system gets closer to the oracle point. Thus with AUC, lower is better.

%[uncomment this text for ArXiv] In order to find the temperature in a reasonable zone between the two anchors, in order to get samples to humans, we defined a completion task, in which the LM is prompted by a prompt of 2sec from a target sentence and has to produce the next units until it reaches another 2sec. We sample 10 completions and compute the BLEU between the completion and the target senteence. A too low temperature produce stereotypical sentences that will miss the target; a too high temperature produce garbage output.  The optimal temperature is used to sample the sentences for human evaluation, and this also gives us a new completion BLEU score to compare models.

To calculate perplexity of the generated utterances, we use a pre-trained ASR\footnote{We use a \textsc{large} wav2vec 2.0~\href{https://github.com/pytorch/fairseq/tree/master/examples/wav2vec\#wav2vec-20}{model}, trained on LibriSpeech-960h with CTC loss from scratch. Its decoder uses the standard KenLM 4-gram language model.} to convert speech to text, and an off-the-shelf Transformer model trained on the English NewsCrawl dataset.\footnote{\href{https://github.com/pytorch/fairseq/tree/master/examples/language_model}{\fontsize{7.5pt}{8pt}\texttt{github.com/pytorch/fairseq/.../language\_model}}}

\subsection{Encoding: Zero-shot probe metrics}\label{sec:bbeval}

The purpose of the encoding metrics is to evaluate the quality of the learned representations at each linguistic level along the pipeline linking the S2u and the uLM. They are inspired by human psycholinguistics and can be  be thought of as \textit{unit tests} providing interpretation and diagnosis. We entirely draw on evaluations from the Zero Resource challenge series \cite{versteegh2016zero,dunbar2019zero,Nguyen2020}\footnote{\fontsize{7.5pt}{7.5pt}\url{www.zerospeech.com}} for comparability with published work and refer to these challenges for details. These metrics are ``zero-shot''  because they do not require training any classifier, and are either based on distances over embeddings, or on computing probabilities over entire utterances. When they have hyperparameters, these are selected using a validation set. 

For acoustic-level evaluation, we use the  between-speaker \textbf{ABX score} to quantify how well-separated phonetic categories are. Briefly, it consists in estimating the probability that two tokens of the same category $A$ ($x$ and $a$) are closer to one another than a token of $A$ ($x$) and of $B$ ($b$). The categories are triphones that only differ in the middle phoneme (like \textit{bit} and \textit{bet}) and the score is averaged over all possible such pairs. For the across-speaker ABX, $a$ and $b$ are spoken by the same speaker and $x$ by a different one, requiring feature invariance over a speaker change. %The distances used in this test are the angular distance frame-wise (arc cos of the normalized dot product), averaged along the dynamic time warping path of the two acoustic tokens. 
We also include the \textbf{bitrate} which has been used in the TTS-without-T challenges \cite{dunbar2019zero} to quantify the efficiency of the discrete units used to resynthetize speech. It is simply the entropy of the sequence of units divided by the total duration. 

For language-level evaluation, we use \textbf{spot-the-word accuracy} from the Zero Resource 2021 Benchmark \cite{Nguyen2020}. It consists in detecting the real word from a pair of short utterances like 'brick' vs 'blick', matched for unigram and bigram phoneme frequency to ensure that low-level cues do not make the task trivial. This task can be done by computing the probability (or pseudo-probability) of the utterances from the uLM. The test set (sWUGGY) consists of 5,000 word-pseudoword pairs generated by the Google TTS API, filtered for the word being present in the LibriSpeech 960h training set \cite{Panayotov2015}. The ZR21 benchmark also uses higher level metrics, notably, syntactic (based on the sBLIMP dataset), which we did not use because the baselines were too close to chance.

\vspace{-0.2em}
\subsection{Human evaluation metrics}\label{sec:heval}
\vspace{-0.4em}
As above, we asked humans to evaluate two aspects of speech generation: intelligibility and meaningfulness. Intelligibility was assessed using two metrics: i) Mean Opinion Scores (MOS) in which raters were asked to evaluate subjectively how intelligible a given audio sample is; ii) Character Error Rate (CER) computed from written transcriptions providing an objective intelligibility test. As for meaningfulness, we set up a meaningfulness-MOS (MMOS) in which raters were asked to evaluate how natural (considering both grammar and meaning) a given sample is. For both subjective tests raters evaluate the samples on a scale of 1-5 with an increment of 1. 

For the MMOS, we had to select a temperature to sample from. Preliminary experiments showed that humans preferred lower temperatures (yielding also less diverse outputs, see Supplementary Section \ref{sup:gene}). Here, we settled on selecting the temperature on a model-by-model basis by constructing a continuation task: we take the 1,000 shortest utterances from LibriSpeech test-clean that are at least 6 seconds long, and use the first 3 seconds as prompts for the uLM (after transcribing them into pseudo-texts). For each prompt, we generated 10 candidate continuations of the same length (in seconds) as the utterance which we took the prompt from. We varied temperature (0.3, 0.4, ..., 1.4, 1.5, 1.7, 1.9, 2.1, 2.3, 2.5, 3.0), and selected the one yielding the maximal BLEU-2 score with the reference sentence (after ASR). These temperatures were typically between the two boundary temperatures described above. 

We evaluated 100 samples from each of the evaluated methods while we enforced at least 15 raters for each sample. The CrowdMOS package~\cite{ribeiro2011crowdmos} was used for all subjective experiments using the recommended recipes for detecting and discarding inaccurate scores. The recordings for the naturalness test were generated by the LM unconditionally and conditionally from  a 3 seconds prompt. Participants were recruited using a crowd-sourcing platform.
%on the Amazon Mechanical Turk Platform.

%%%%%%% OVERALL FIGURE %%%%%%%%
\begin{figure*}
    \centering
    \includegraphics[width=0.95\textwidth]{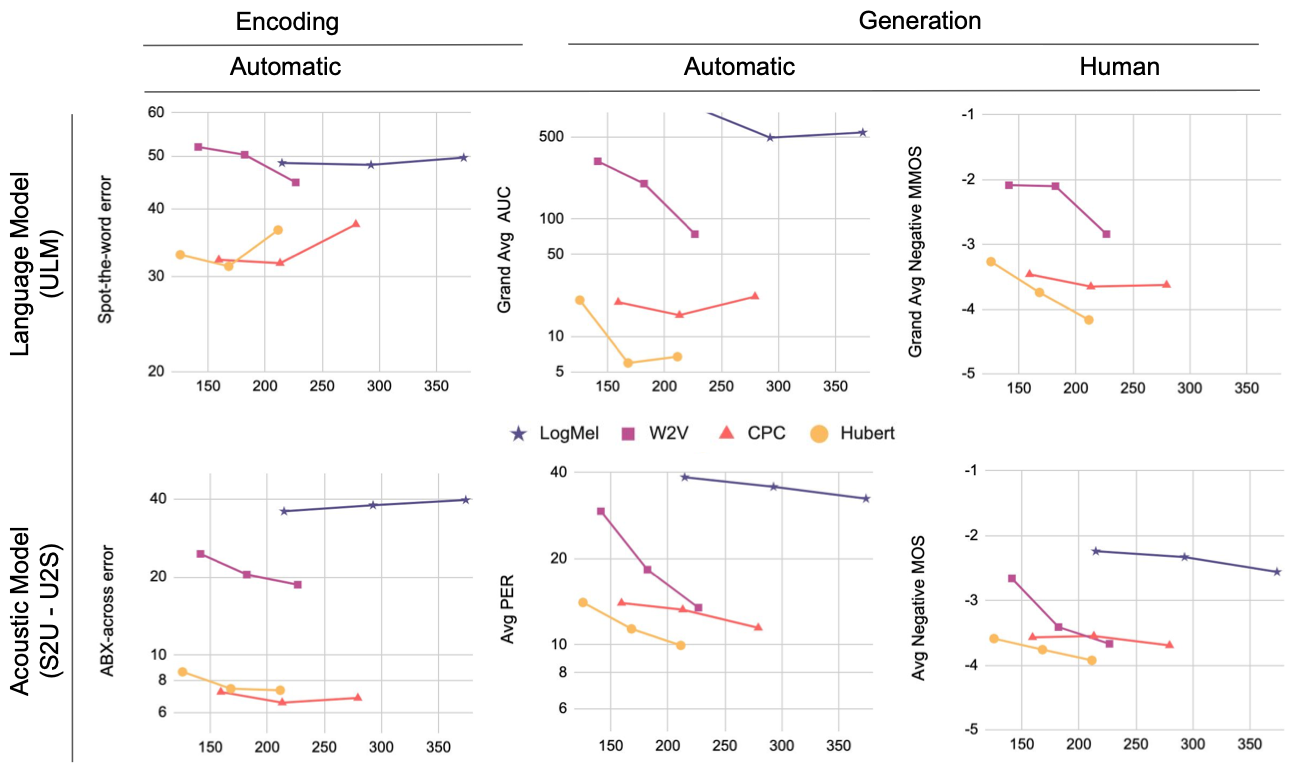}

    \vspace{-0.7em}
    \caption{\textbf{Overall results with automatic and human metrics}.  The results are presented in terms of bitrate for 4 encoders (LogMel, CPC, HuBERT and wav2vec 2.0) varying in number of units (50, 100, 200). For definition of the tasks and metrics, see Table~\ref{tab:metric_overview} and Figure~\ref{fig:setup}. Negative human opinion scores are shown for ease of comparison with automatic metrics (lower is better). The generation metrics have been averaged across LS and LJ (PER and MOS; resynthesis task) and across prompted and unprompted conditions (AUC and MMOS; speech generation task). The Log Mel Fbank based systems were not evaluated by humans in the speech generation task.}
    \label{fig:overall}
    \vspace{-0.7em}
\end{figure*}
%%%%%%% OVERALL FIGURE %%%%%%%%

% \vspace{-.5em}
\section{Proposed Systems}\label{sec:models}
% \vspace{-.2em}

Here, we present our S2u (Section \ref{sec:encoders}), uLM (Section \ref{sec:LMs}) and u2S (Section \ref{sec:pTTS}) components.

\subsection{Speech-to-unit Models}\label{sec:encoders}
We selected 3 recent state-of-the-art unsupervised Encoders, which we used 'out of the box': we did not retrain them nor change their hyperparameters. We also included a log Mel filter-bank baseline (80 filters, computed every 10ms). We then discretized the embeddings using k-means. We only give a high level description of these models, and refer to the original publications for details. % {\color{red} Section} \ref{sec:sumodels}. 

\noindent{\textbf{CPC.}} Contrastive Predictive Coding \cite{van2017neural} as applied to speech consists of two components: an encoder and a predictor. The encoder produces an embedding $z$ from speech input. The predictor predicts the future states of the encoder based on the past, and the system is trained with a contrastive loss. We use the CPC model from \cite{Riviere2020}, which was trained on a ``clean'' 6k hour sub-sample of the LibriLight dataset~\cite{Kahn2020,Riviere2020}. We extract a representation from an intermediate layer of the predictor, which provides a 256-dimensional embedding (one per 10ms), as in the original paper.%based on the ABX score on Libri-light dev set. 

\noindent{\textbf{wav2vec 2.0.}} Similar to CPC, this model uses an encoder and a predictor, which is trained contrastively to distinguish positive and negative samples from discretized and masked segments of the encoder's output. We use the \textsc{large} variant of pretrained wav2vec 2.0 \cite{baevski2020wav2vec} trained on 60k hours of LibriLight dataset \cite{Kahn2020}. This model encodes raw audio into frames of 1024-dimensional vectors (one per 20ms). To choose the best layer, we extracted frozen representations of the 10 hour LibriLight subset from every layer of the model and trained a linear classifier with the CTC loss to predict the phonetic version of the text labels. %We use G2P \cite{g2pE2019} phonemizer to obtain 71 class phoneme labels from the original text transcriptions. 
Layer 14 obtained the lowest PER on LS dev-other (a similar approach was done in \cite{baevski21} which in this case selected Layer 15).

\noindent{\textbf{HuBERT.}} Unlike CPC and wav2vec 2.0 that use a contrastive loss, HuBERT is trained with a masked prediction task similar to BERT~\cite{devlin-etal-2019-bert} but with masked continuous audio signals as inputs. The targets are obtained through unsupervised clustering of raw speech features or learned features from earlier iterations, motivated by DeepCluster~\cite{caron18deepcluster}. We use the \textsc{Base} 12 transformer-layer model trained for two iterations~\cite{hsu2020hubert} on 960 hours of LibriSpeech~\cite{Panayotov2015}. This model encodes raw audio into frames of 768-dimensional vectors (one per 20ms) at each layer and we extract those from the 6$^{th}$ layer as in the original paper.

\noindent{\textbf{LogMel.}} As a baseline, we consider a Log Mel Filterbank encoder using 80 frequency bands. %One u2S model is trained for each unit type (e.g., CPC-50) as described in \ref{sec:pTTS}.% %%%and \ref{sup:pTTS}. 

% \paragraph{VQ-VAE.}  Here we use the Vector Quantized Variational Auto Encoder (VQ-VAE) model as introduced by~\citet{van2017neural}. VQ-VAE model performs in a similar manner to a Variational Auto Encoder~\cite{kingma2013auto} where the output of the encoder is discrete units rather than continuous values. During training we keep the prior distribution over the discrete latent units constant and uniform. Similarly to CPC models, we trained VQ-VAE models on the ``clearer'' 6K hours subset from LibriLight dataset. 
% %\dpx{@yossi/wei-ning, please add some details}

\noindent{\textbf{Quantization.}} We use k-means to convert continuous frame representations into discrete representation by training on LibriSpeech clean-100h \cite{Panayotov2015}. We experiment with codebooks that have 50, 100, and 200 units.

\subsection{unit-Language Model}\label{sec:LMs}
We use the Transformer model as implemented in \texttt{fairseq}~\cite{Ott2019}. We use the \textit{transformer\_lm\_big} architecture: it has 12 layers, 16 attention heads, embedding size of 1024, FFN size of 4096, and dropout probability of 0.1, and we train it as a causal LM on sequences of pseudo-text units. Each sample contains up to 3,072 units. 
% Language generation is done by sampling with temperature.
We use sampling with temperature for generation.
% Language generation is done by sampling with temperature.

All language models are trained on ``clean'' 6k hours sub-sample of LibriLight used in~\cite{Riviere2020}, transcribed with corresponding discrete units.
In preliminary experiments, we found that removing sequential repetitions of units improves performance, hence we apply it universally.\footnote{For example, a pseudo-text \texttt{10 11 11 11 21 32 32 32 21} becomes \texttt{10 11 21 32 21}.} We hypothesise that this simple modification allows to use Transformer's limited attention span more efficiently as in  \citet{hsu2020text}.

%\dpx{@euqgene, says about the training set}

\subsection{unit-To-Speech Model}\label{sec:pTTS}
% We use \emph{Tacotron} model~\cite{shen2018natural} which gets as input the discrete units and outputs a log mel-spectrogram. 
We adapt the \emph{Tacotron-2} model~\cite{shen2018natural} such that it takes pseudo-text units as input and outputs a log Mel spectrogram. To enable the model to synthesize arbitrary unit sequences, including those representing incomplete sentences, we introduce two modifications. First, we append a special ``end-of-input'' (EOI) token to the input sequence, hinting the decoder to predict the ``end-of-output'' token when attending to this new token. However, this modification alone may not be sufficient, as the decoder could still learn to ignore the EOI token and correlate end-of-output prediction with the learned discrete token that represents silence as most of the speech contains trailing silence. To address this, we train the model using random chunks of aligned unit sequence and spectrogram, and append the EOI token to unit sequence chunks, such that the audio does not always end with silence. We implement chunking in the curriculum learning fashion, where the chunk size gradually grows (starting with 50 frames with an increment of 5 per epoch) to increase the difficulty of the task. For waveform generation, we use the pre-trained flow-based neural vocoder \emph{WaveGlow} ~\cite{prenger2019waveglow}. This model outputs the time-domain signal given the log Mel spectrogram as input.
% This model gets as input the log Mel spectrogram and outputs the time-domain signal.
All u2S models were trained on LJ Speech (LJ)~\cite{ljspeech17}.

%%%%%%%%%%%%%%%%%%%%%%%%%%%%%%%%%%%%%%%%%%%%%%%%%%%%%%%%%%%%%%%%
%%%%%%%%%%%%%%%%%  RESULTS %%%%%%%%%%%%%%%%%%%%%%%%%%%%%%%%%%%%%
%%%%%%%%%%%%%%%%%%%%%%%%%%%%%%%%%%%%%%%%%%%%%%%%%%%%%%%%%%%%%%%%
\newcommand{\mc}[3]{\multicolumn{#1}{#2}{#3}}
\newcommand{\ux}[1]{\underline{#1}}
\newcommand{\hs}[1]{\hspace{#1\tabcolsep}}
\newcommand{\da}[0]{$\downarrow$}
\newcommand{\ua}[0]{$\uparrow$}
% \vspace{-.5em}

\vspace{-.2em}

\section{Results}\label{sec:results}
% The main results are shown in Table \ref{tab:main}. 
\vspace{-.2em}

In Figure \ref{fig:overall}, we report the overall results of our models and our LogMel baseline as a function of the number of quantized units on our main automated and human metrics. More detailed results follow in the following sections, including two character-based toplines: one uses the oracle transcripts for training the LM, the other uses transcripts produced by the pre-trained ASR model.% We also consider one baseline based on k-means over Mel Filterbanks. %One u2S model is trained for each unit type (e.g., CPC-50) as described in \ref{sec:pTTS}.% %%%and \ref{sup:pTTS}. 

%%%%%%%%%%%%%%%%%%%%%%%%%%%%%%%%%%%%%%%%%%%%%%%%%%%%%%%%%%%%%%%%%%%%%%%%%%%
%%%%%%%%%%%%%%%%%%%%%% RESYNTH RESULTS TABLE BEGIN %%%%%%%%%%%%%%%%%%%%%%%%
%%%%%%%%%%%%%%%%%%%%%%%%%%%%%%%%%%%%%%%%%%%%%%%%%%%%%%%%%%%%%%%%%%%%%%%%%%%

 \begin{table*}[t]
 \caption{\textbf{Results on the resynthesis task} for 3 unsupervised models plus one LogMel baseline and 3 unit sizes. Bitrates are in bit/sec, PER are for a pretrained phone recognition model without lexicon and LM, CER are derived from a full ASR model (lower is better). Human MOS (upper is better) and CER (computed from transcription, lower is better) are provided (the 95\% confidence interval was on average .32 for MOS and 1.8 for human CER)}\label{tab:resynth}
 \centering
 \resizebox{\textwidth}{!}{
 \begin{tabular}{l c c|| c c | c c | c c | c c }
 \hline
 \mc{3}{l||}{Systems}  & \mc{4}{c|}{End-to-end ASR-based metrics}       &\mc{4}{c}{Human Opinion}\\
 \hline
 S2u                                             & Nb    &Bit-  & PER\da  & PER\da  & CER\da & CER\da  & MOS\ua & MOS\ua & CER\da& CER\da    \\ %& Bit-     
 architect.                                      & units &rate  & (LJ)    & (LS)    & (LJ)   & (LS)    & (LJ)   & (LS)   & (LJ)  &(LS)   \\%& rate     
 \hline
 \mc{2}{l@{\hs{0.1}}}{\textit{Toplines}}               &      &         &         &        &         &        &       &        & \\%&  
 %\mc{2}{l@{\hs{0.1}}}{oracle text}                    &      &-        &-        &-       &-        & 4.28   &       &        & \\%& -        
 \mc{2}{l@{\hs{0.1}}}{original wav}                    &      &-        &-        &-       &-        & 4.83   &  4.30 &  8.88  &  6.73\\%&-         
 \mc{2}{l@{\hs{0.1}}}{orig text+TTS}                   &      & 7.78    & 7.92    & 8.87   & 5.14    & 4.02   &  4.03 &  13.25 &  10.73\\%&-         
 %\mc{2}{l@{\hs{0.1}}}{\textit{Supervised}}            &      &         &         &        &         &        &       &        & \\%&         
 ASR + TTS                                       & 27    &      & 9.45    & 8.18    &   9.48 &   5.30  &  4.04  & 4.06  &  15.98 & 11.56\\%& -        
 \hline
 \mc{2}{l@{\hs{0.1}}}{\textit{Baselines}}                &      &         &         &        &         &        &       &            & \\%&          
 LogMel                                          & 50    & 214.8&   27.72 &   49.38 &  27.73 &  52.05  &   2.41 & 2.07  &   43.78    & 66.75\\%& 214.8    
 LogMel                                          & 100   & 292.7&   25.83 &   45.58 &  24.88 &  48.71  &   2.65 & 2.01  &   37.39    & 62.72\\%& 292.7    
 LogMel                                          & 200   & 373.8&   19.78 &   45.16 &  17.86 &  46.12  &   2.96 & 2.16  &   23.33    & 62.6\\%& 373.8    
 \hline
 \mc{2}{l@{\hs{0.1}}}{\textit{Unsupervised}}             &      &         &         &        &         &        &       &            & \\%&          
 CPC                                             & 50    & 159.4&   10.87 &   17.16 &  10.68 &  12.06  &   3.63 & 3.51  &    13.97   & 19.92\\%& 159.4    
 CPC                                             & 100   & 213.1&   10.75 &   15.82 &   9.84 &   9.46  &   3.42 & 3.68  &    13.53   & 14.73\\%& 213.1    
 CPC                                             & 200   & 279.4&\bf 8.74 &   14.23 &   9.20 &   8.29  &   3.85 & 3.54  &  \bf9.36   & 14.33\\%& 279.4    
 HuBERT-L6                                       & 50    & 125.7&   11.45 &   16.68 &  11.02 &  11.85  &   3.69 & 3.49  &    14.54   & 13.14\\%& 125.7    
 HuBERT-L6                                       & 100   & 168.1&   9.53  &   13.24 &   9.31 &   7.19  &   3.84 & 3.68  &    13.02   & 11.43\\%& 168.1    
 HuBERT-L6                                       & 200   & 211.3&   8.87  &\bf11.06 &\bf8.88 & \bf5.35 &\bf4.00 &\bf3.85&    11.67   & \bf10.84\\%& 211.3    
 wav2vec-L14                                     & 50    & 141.3&  24.95  &   33.69 &  25.42 &  32.91  &   2.45 & 2.87  &    46.82   & 54.9\\%& 141.3    
 wav2vec-L14                                     & 100   & 182.1&  14.58  &   22.07 &  13.72 &  17.22  &   3.50 & 3.32  &    23.76   & 28.1\\%& 182.1    
 wav2vec-L14                                     & 200   & 226.8 &  10.65  &   16.34 &  10.21 &  10.50  &   3.83 & 3.51  &    13.14   & 15.27\\%& 226.8    
 %VQ-VAE     & 50     &         &         &        &          &          &         &         &        &        &         &         &        &          &         &        &        &       &          \\
 %VQ-VAE     & 100    &         &         &        &          &          &         &         &        &        &         &         &        &          &         &        &        &       &          \\
 %VQ-VAE     & 200    &         &         &        &          &          &         &         &        &        &         &         &        &          &         &        &        &       &          \\
 \hline
 \end{tabular}
 }
 \end{table*}

%%%%%%%%%%%%%%%%%%%%%%%%%%%%%%%%%%%%%%%%%%%%%%%%%%%%%%%%%%%%%%%%%%%%%%%%%%%
%%%%%%%%%%%%%%%%%%%%%%%%%  RESYNTH TASK TABLE END %%%%%%%%%%%%%%%%%%%%%%%%%
%%%%%%%%%%%%%%%%%%%%%%%%%%%%%%%%%%%%%%%%%%%%%%%%%%%%%%%%%%%%%%%%%%%%%%%%%%%

\vspace{-.2em}

\subsection{Results on the resynthesis task}
\vspace{-.2em}

Overall resynthesis results are shown in the bottom middle and right cells of Figure~\ref{fig:overall} for our main automatic (PER) and human scores (MOS), respectively, averaged across the LS and LJ evaluation sets. We observe that across all models, increasing the number of units uniformly leads to better scores suggesting that the u2S component can take benefit from extra details of the input to produce a more realistic output. HuBERT and CPC seem to be giving the best results, for both humans and models better capturing phonetic information than other models at equivalent bitrates.

% The best models were CPC and HuBERT with a PER of 8.7\% and 8.9\% respectively, intermediate between the ASR+TTS (9.45\% error) and the original text+TTS (7.78\%) toplines. The human MOS was best for HuBERT (4.00) and approached both toplines (4.02 and 4.04, resp.) and the CER was best for CPC (9.36\%) outperforming the toplines (15.98\% and 13.25\%, resp.). These scores showed excellent resynthesis scores approaching that of the original waveform (MOS 4.83; CER: 8.88\%).

More detailed results are in Table \ref{tab:resynth} separating the scores for the LJ and LS resynthesis, and adding extra automatic metrics (CER) and human metrics (human CER). On PER, we found a domain effect: resynthesizing input from LJ Speech yields lower PER than from LibriSpeech on all unsupervised models. From the viewpoint of the encoder, LJ Speech is out-of-domain; therefore, one would expect that the units are making more errors than for the trained LibriSpeech. On the other hand, the u2S component has learned from LJ Speech encoded with these units, and might have learned to compensate for these lower quality units. When LibriSpeech is offered as input, the u2S component cannot adapt to this nominally better input and ends up yielding lower quality outputs. This observation is worth further explorations, as other metrics like CER (using an LM) and human evaluations only replicated this for the models with the lowest score (like LogMel and wav2vec). The automatic PER and CER scores and the human MOS and CER scores, all correlate well with one another across the $4 \times 3$ models and baselines. Within the LJ or LS domain , the Pearson r ranged from .95 to .99; across domains it was less good (from .79 to .96) illustrating again the existence of a domain effect. Not shown here, we reached similar conclusions with our fitted-ASR metrics, but with less good scores and correlations.  Table \ref{tab:resynth} also shows the results of the two toplines (original text+TTS and ASR+TTS). Interestingly, our best models come within 3\% absolute in PER or CER compared to these toplines, are quite close to them in terms of MOS and even beat them in terms of human CER.

\vspace{-0.2em}
\subsection{Results on the generation task}
\vspace{-0.2em}

 \begin{table*}[t]
 \caption{\textbf{Results on the generation task} for three unsupervised models plus the LogMel baseline and 3 unit sizes. PPX@-o-VERT and VERT@o-PPX are reported as PPX and VERT. '-' : missing or non calculable results. Human MMOS are also provided (the 95\% confidence interval was on average .29 for uncond. and .61 for cond.).}
 \label{tab:gene}
 \centering
 \resizebox{\textwidth}{!}{
 \begin{tabular}{l c || c c  c | c  c c | c c}
 \hline
 \mc{2}{l||}{Systems}  & \mc{6}{c|}{Generation based metrics}       &\mc{2}{c}{Human Opinion}\\
 \hline
 Encoder                                         & Nb     &\mc{3}{c|}{\ux{unconditional}}       &\mc{3}{c|}{\ux{prompt}}    &  \ux{uncond}. & \ux{prompt} \\
 architect.                                      & units  & PPX\da    & VERT\da &  AUC\da  & PPX\da   &VERT\da  & AUC\da    & MMOS\ua & MMOS\ua     \\
 \hline
 \mc{2}{l@{\hs{0.1}}||}{\textit{Controls}}                &           &         &          &          &         &           &                 \\
 \mc{2}{l@{\hs{0.1}}||}{oracle text}                      & 154.5     &  19.43  &     -    &   154.5  &  19.43  &   -       & 4.02  &  4.26   \\
 \mc{2}{l@{\hs{0.1}}||}{ASR + LM}                         & 178.4     & 21.31   &    0.18  &    162.8 &  20.49  &     0.04  & 3.91  &  4.38   \\
 \hline
 \mc{2}{l@{\hs{0.1}}||}{\textit{Baseline}}                &           &         &          &          &         &           &           \\
 LogMel                                          & 50     &  1588.97 &    -    &  1083.76   &  -       &   -     &      -    &   -    & -      \\
 LogMel                                          & 100    &  1500.11 &   95.50  & 510.26   &  -       &   -     &     -     &   -    & -      \\
 LogMel                                          & 200    &  1539.00  &     -   &  584.16     &  -       &   -     &     -     &   -    & -      \\
 \hline
 \mc{2}{l@{\hs{0.1}}||}{\textit{Unsupervised}}            &           &         &          &          &         &          &          \\
 CPC                                             & 50     &   374.26   &   46.26 &  19.68   &  323.9   & 39.92   &   18.44  &  3.31  & 3.61   \\
 CPC                                             & 100    &   349.56   &   41.797 & 15.74  &  294.7   & 42.93   &   14.06  &  3.65  & 3.65   \\
 CPC                                             & 200    &  362.84   &   40.28 &   16.46  &   303.5  & 43.42   &   26.67  &  3.58  & 3.67   \\
 HuBERT-L6                                       & 50     &  376.33  &   43.06 &   19.27 &   339.8  &  45.85  &   21.03  &  3.53  & 3.00    \\
 HuBERT-L6                                       & 100    &\bf 273.86  &\bf 31.36 &\bf 5.54  &\bf 251.2 &\bf 33.67&\bf 5.88  &  3.95  & 3.53    \\
 HuBERT-L6                                       & 200    &  289.36   &   33.04 &   7.49   &   262.4  &   34.30 &     6.13 &\bf 4.01 &\bf 4.32  \\
 wav2vec-L14                                     & 50     &   936.97   &    -    &  307.91  &   1106.3 &      -  &    330.8  &  2.26 & 1.91    \\
 wav2vec-L14                                     & 100    &   948.96  &  79.51 &   208.38 &   775.1  &      -  &    205.7  &  2.28 & 1.92    \\
 wav2vec-L14                                     & 200    &   538.56   & 61.06  &  61.48 &   585.8  &      -  &    91.07  &  2.64 & 3.04    \\
\hline
 \end{tabular}
 }
 \end{table*}
%%%%%%%%%%%%%%%%%%%%%%%%%%%%%%%%%%%%%%%%%%%%%%%%%%%%%%%%%%%%%%%%%%%%%%%%%%%
%%%%%%%%%%%%%%%%%%% GENERATION RESULTS NEW TABLE END %%%%%%%%%%%%%%%%%%%%%%%%%%
%%%%%%%%%%%%%%%%%%%%%%%%%%%%%%%%%%%%%%%%%%%%%%%%%%%%%%%%%%%%%%%%%%%%%%%%%%%

The upper mid and right cells of Figure~\ref{fig:overall} shows generation results averaging across the unconditional and conditional conditions, on automatic and human evaluations respectively. 
The main result is that there is both an effect of number of units and of system. As for resynthesis, 50 units is always worst, but contrary to resynthesis, 200 units is not always better. Overall, the results on generation are congruent with the idea that speech generation both requires good scores on language modeling and on speech synthesis. The best results for a particular model are then a compromise between the number of units that give both scores to either of these tasks. In terms of systems, the best one here is HuBERT. 
Regarding human evaluations, they show similar patterns with a clear dispreference for 50 units, and either 100 or 200 being better. 
% Here HuBERT come as the best systems. 

Detailed results are shown in Table \ref{tab:gene} with separate statistics for conditional and unconditional generation and additional results with PPX@o-VERT and VERT@o-PPX. As expected, the perplexity metric improved with prompts, but not the diversity score. The human results are congruent with the automatic scores, although they tend to prefer more units, perhaps showing that they cannot fully dissociate their judgment of meaning from their judgment of intelligibility. The three metrics correlate well with one another (r between .86 and .99) and correlate with their counterpart across task (prompted vs.\ unprompted: r between .82 and .99). Human evaluations correlated well with the automatic metrics (AUC: r=.87; PPX: r=.92; VERT: r=0.75). %A sub-analysis reveals that lower temperature (temp@ppx) generations were almost always preferred than the other two sampling points. Samples from this temperature show the same pattern of results as the average across the three temperatures (see Supplementary Section \ref{sup:gene} for details).
\vspace{-.1em}
\subsection{Results for zero-shot probe metrics}
\vspace{-.1em}

 %%%%%%%%%%%%%%%%%%%%%%%%%%%%%%%%%%%%%%%%%%%%%%%%%%%%%%%%%%%%%%%%%%%%%%%%%%%
 %%%%%%%%%%%%%%%%%%% ZERO SHOT RESULTS TABLE BEGIN %%%%%%%%%%%%%%%%%%%%%%%%
 %%%%%%%%%%%%%%%%%%%%%%%%%%%%%%%%%%%%%%%%%%%%%%%%%%%%%%%%%%%%%%%%%%%%%%%%%%%
 \begin{table}[t!]
 \caption{\textbf{Results for zero-shot probe metrics} for 3 unsupervised models plus one LogMel baseline and 3 unit sizes. ABX within and across speakers, spot-the-word and acceptability judgments are error rates (lower is better); chance is 50\%.}
 \resizebox{\columnwidth}{!}{
 \label{tab:zero}
 \begin{tabular}{l @{\hs{0.4}}c@{\hs{0.4}} || c @{\hs{0.8}}c | c@{\hs{0.8}} c }%| c c c c | c c c c c | c c c}
 \hline
 \mc{2}{r||}{Metrics} & \mc{2}{c|}{S2u} &\mc{2}{c}{uLM}\\
 %\hline
 \cline{2-6}
              & Nb    & ABX    & ABX     & spot-the-& accept.   \\%& Bit-    
 System       & units &with.\da& acr.\da & word\da  & judg.\da   \\%& rate    
 \hline
 \mc{2}{l@{\hs{0.1}}||}{\textit{Toplines}}&&      &          &           \\
 %\mc{2}{l@{\hs{0.1}}||}{\textit{Controls}}&&&      &          &           \\
 %\mc{2}{l@{\hs{0.1}}||}{oracle text}&-&-&-         &-         &-          \\
 %\mc{2}{l@{\hs{0.1}}||}{original wav}&-&-&-        &-         &-          \\
 % \mc{2}{l@{\hs{0.1}}||}{oracle char+LM}&-&-&-      &     &      \\
 \mc{2}{l@{\hs{0.1}}||}{ASR+LM}&-&-      &    3.12 & 29.02  \\
 %\hline
 %\mc{2}{l@{\hs{0.1}}||}{\textit{Supervised}}&&&    &          &           \\
 %Oracle ASR  & 27?    &        &         &        &          &           \\
 % ASR + LM+ TTS & 27   & -       & -       & -      & -        & -         \\
 \hline
 \mc{2}{l@{\hs{0.1}}||}{\textit{Baselines}}&&     &          &           \\
 LogMel      & 50     &   23.95 &  35.86 &   48.52  &  46.78       \\%& 214.8   
 LogMel      & 100    &   24.33 &  37.86 &   48.12  &  46.83         \\%& 292.7   
 LogMel      & 200    &   25.71 &  39.65 &   49.62  &  47.76     \\%& 373.8   
 \hline
 \mc{2}{l@{\hs{0.1}}||}{\textit{Unsupervised}}&&  &          &           \\%
 CPC         & 50     &    5.50 &   7.20 &    32.18 &    45.43  \\%& 159.4   
 CPC         & 100    &\bf 5.09 &\bf6.55 &    31.72 &    44.35  \\%& 213.1   
 CPC         & 200    &    5.18 &   6.83 &    37.40 &    45.19  \\%& 279.4   
 
 HuBERT-L6   & 50     &    7.37 &   8.61 &    32.88 &	  44.06  \\%& 125.7   
 HuBERT-L6   & 100    &    6.00 &   7.41 &\bf 31.30 &\bf 42.94  \\%& 168.1   
 HuBERT-L6   & 200    &    5.99 &   7.31 &    36.52 &	  47.03  \\%& 211.3   
 wav2vec-L14 & 50     &   22.30 &  24.56 &    51.92 &    45.75  \\%& 141.3   
 wav2vec-L14 & 100    &   18.16 &  20.44 &    50.24 &    45.97  \\%& 182.1   
 wav2vec-L14 & 200    &   16.59 &  18.69 &    44.68 &    45.70  \\%& 226.8   
 %VQ-VAE     & 50     &         &         &        &          &               \\
 %VQ-VAE     & 100    &         &         &        &          &               \\
 %VQ-VAE     & 200    &         &         &        &          &               \\
 \hline
 \end{tabular}}
 \vspace{-.7em}
 \end{table}
%%%%%%%%%%%%%%%%%%%%%%%%%%%%%%%%%%%%%%%%%%%%%%%%%%%%%%%%%%%%%%%%%%%%%%%%%%%
 %%%%%%%%%%%%%%%%%%% ZERO SHOT RESULTS TABLE END %%%%%%%%%%%%%%%%%%%%%%%%
 %%%%%%%%%%%%%%%%%%%%%%%%%%%%%%%%%%%%%%%%%%%%%%%%%%%%%%%%%%%%%%%%%%%%%%%%%%%
 
In Table \ref{tab:zero}, we show the results for zero-shot metrics across the different models and baselines. Overall, the performances depend on the linguistic levels while remaining above chance. While performances are excellent at the acoustic level (6.5\% error for the best model on ABX-across), they are intermediate at the lexical level (31.3\% error for the best model on spot-the-word). Not shown, the syntactic test is close to chance (42\% error for the best model on the sBLIMP test). These values are worse than the ASR-topline (3.1\% and 29\%, for lexicon and syntax resp.), showing room for improvement.

The metrics correlate well: the ABX score predicts the lexical score ($r=0.85$) and the syntax score ($r=0.71$). Across the different models, CPC gets the best units (ABX score) and HuBERT gets the best LM scores. In addition, we see a clear effect of number of units (Figure~\ref{fig:overall}). For wav2vec, the performances on all metrics increase with more units, whereas, for CPC and HuBERT a U-shaped pattern emerges on most metrics, with best scores for units of intermediate sizes. It is interesting that the models with the highest bitrate do not always have the best results. This means that encoding too much acoustic information can be detrimental to linguistic encoding in the uLM. See Appendix Section \ref{sup:zero} showing that ABX has good correlations with automatic and human metrics ($r>.88$).%[optional]There appears to be a sweetspot around 200 bit/second. 
%Here, we compare the automatic ASR metrics with the human evaluation. We find that MOS correlates with an R of xx with the ASR-CER. Refer to the gigantic correlation table in the Supplementary. 

%Add here the correlations of ABX and others on the automatic and human metrics.

\vspace{-.2em}
\section{Discussion and Conclusion}\label{sec:discuss}
\vspace{-.2em}

We introduced \textit{Generative Spoken Language Modeling} as a new unsupervised task bridging the gap between speech and natural language processing and related it conceptually to previously studied unsupervised tasks: Acoustic Unit Discovery, Spoken Language Modeling, Discrete Speech Resynthesis and Text Generation. 
We introduced a suite of metrics, baselines, and first results on Librilight that sets the playing field for future work. For comparability, we open source our evaluation stack and the best of our baseline models.

%Like the first three tasks, Speech Generation needs to learn good speech representations in an unsupervised way, like the second and fourth task, it needs to learn higher order linguistic representations as well, and like the fourth task new utterances can be sampled from the model, conditionally or unconditionally. What is new, is that through the combination of these capability, we end up with a model learning to produce novel speech utterances in a completely unsupervised way, without any textual or categorical labels. 

Our main contributions are as follows: (1) we established a set of easy to use automatic ASR-based metrics for model comparison at two critical levels for this task: intelligibility of the speech output and meaningfulness in terms of higher linguistic content. We assessed the first through ASR-based PER and CER metrics; and the second using text-generation-based metrics (AUC for PPX/VERT). (2) We found that these two sets of metrics correlated well with human judgement and (3) that they can be approximated with their inference-mode counterparts, which are faster to computed using zero-shot probe tasks. (4) Applying these metrics to pipeline models based on current speech representation learning models and out-of-the-box LM and TTS components, we found that our basic premise is fulfilled: it is possible to train a language model from quantized units derived from audio and using it to generate new speech. The generated speech is English sounding, with recognizable phonemes and words and locally acceptable syntax (see transcribed examples in the Appendix and audio snippets here: {\footnotesize\url{\urldemo}}). Our automatic metrics confirm the quality of the representations and outputs at the acoustic/phonetic level, but show that improvements are needed at the language level. %: the lexical and syntactic zero-shot tests show performance below that of text-based models, and the generations are not plausible beyond a few seconds. 
It is to be expected that performance will increase with larger training sets beyond our 6k hours, as has been noted in the case of text. (5) We also uncovered specific issues regarding the number of quantized units. For speech resynthesis, the optimum number of units was always 200 by a large margin, reflecting the well known bitrate/intelligibility trade-off \cite{dunbar2019zero}. However, for language modeling, this was not necessarily the case, as the more detailed acoustic information may introduce too numerous phonetic details that have no impact at the level of lexical and syntactic representations. (6) Finally, we found that the choice of units also affected the temperature parameter which is used to control the trade-off between quality and diversity in text-based language model. To address this effect, we proposed a method to normalize the temperature by using an oracle text to build perplexity and diversity anchor points.

\setcounter{figure}{0}
\setcounter{table}{0}
%%\setcounter{page}{1}
%\makeatletter
%\renewcommand{\theequation}{A\arabic{equation}}
\renewcommand{\thefigure}{A\arabic{figure}}
\renewcommand{\thetable}{A\arabic{table}}
%\renewcommand{\bibnumfmt}[1]{[A#1]}
%\renewcommand{\citenumfont}[1]{A#1}
%%%%%%%%%%%%%%%%%%%%%%%%%%%%%%%%%%%%%%%%%%%%%%%%%%

\begin{table*}[]
 %\footnotesize
 \caption{Unconditional uLM (trained on CPC-100 units) samples, transcribed by ASR, at different temperatures.}\label{tab:samples}
  \resizebox{\textwidth}{!}{
 \begin{tabular}{lp{20cm}}
 \hline
 Temp&Example\\
 \hline
 \multicolumn{2}{l}{\textit{Very low temperature samples  (stuttering zone)}}\\% : low within \& between utterance diversity, low perplexity} \\
 0.3 & \texttt{the property by james resell red for liberata or by jason downy the property by jason downy the property the property the property the property} \\
 0.3 & \texttt{and to take in another path and to take in another path and to take in another path and to take in another path and to take in another path and to take in another path and take in a}\\
 \multicolumn{2}{l}{\textit{Low temperature samples (obsessive zone)}}\\%: high within, low between utt. diversity, mid perplexity} \\
 0.7 & \texttt{chapter nineteen of the life of the upper part of the ocean this is ali bravos recording only bravos recordings are in the public domain i for more information or to volunteer} \\
 0.7 & \texttt{this is a lipper vox are courting oliver vox or courting are in the public domain for afraid art to volunteer pleases it lipper vox dot or this } \\
 \multicolumn{2}{l}{\textit{Mid temperature samples}}\\%: high within, med between utt. diversity, high perplexity} \\
 1.0 & \texttt{but it is attendant from the people to defend himself from this information pride of the potential in criminal activity a curiosity and impetuosity of the world a war soon acquired}\\
 1.0 & \texttt{finally we ought to have a strong plan a without positively the best type of the public with which we ascend it or extend it our business and as we are a persons of the most strong designs and other affairs of the case we }\\
 \multicolumn{2}{l}{\textit{High temperature samples  (babble zone)}}\\%: high within \& between utt. diversity, very high perplexity} \\
 1.5 & \texttt{ation of pure blue he said at once a licking streamy at her warm spot of half performed note was a raging oath let it as bir of amole in mood strolling er crass} \\
 1.5 & \texttt{at the swing here as to motions out of the events not time and abe he was any stump headed and flow any he's the kiln are tama why do ye take the floor} \\
 \hline
 \end{tabular}}
 \vspace{-1em}
 \end{table*}

Obviously, this is only a first step towards building textless NLP applications that could be applied to any language, even low resource ones. To reach this long term goal, three important challenges need to be addressed. 

First, even though we did compare three different encoders and obtained different results we cannot conclude that one encoder is definitely superior to the others. Our point here was merely to use previously published pretrained encoders, and study systematically the effect of number of units on these encoders. A fuller study including a wider set of encoders and a proper hyperparameter search (including the selection of the embedding layer and the clustering algorithm) would be needed in order to determine which of them is most appropriate for speech generation. 

%This was motivated because while LMs and speech synthesis systems are well studied and it was easy to select good out-of-the-box systems, unsupervised encoders are less well known and there was no accepted state-of-the-art. In fact, one additional contribution of this study is a new way to evaluate unsupervised encoders with a downstream (unsupervised) application. 

Second, it is to be expected that to further improve generation results, more needs to be done than applying this pipeline to larger training sets.  Contrary to text, speech unfolds through time and varies continuously in phonetic space. Speech also contains multilayered representations (phonetic, prosodic, speaker identity, emotions, background noise, etc.). However, both our TTS and our LM were out-of-the-box systems typically used for text applications. More work is needed to adapt these architectures to the richness and variability of the speech signal (see \citeauthor{polyak2021speech}, \citeyear{polyak2021speech}, for first steps towards integrating prosody into discrete units). The metrics and baselines we introduced here provide landmarks against which we will measure future progress. 

Third, the automatic metrics that we defined here depend on textual resources to build the evaluation ASR and LM models, and on linguistic resources to build the zero-shot  metrics. How could this ever be applied to low-resource languages? Note that the linguistic resources we require are used only for model selection, not model training. Our metrics allow for fast iterations in architecture and hyperparameter search, but the overall algorithm is totally unsupervised. Therefore, an important next step is to extend this work to other languages, in order to find a common architecture/hyperparameter set that gives good results in held out languages (high or low resource). %We know that this is achievable, because human babies are not preprogrammed to learn only the language of their genetic parents, showing that there exists a generic architecture able to learn any language in the environment. Therefore,
The hope is that once good learning models are tuned using a diverse sample of high resource languages, the same models could be deployed in languages where no such resources are available, and work in a purely unsupervised fashion.

%however their introduction is only to enable fast iterations for architecture/hyperparameter search. O 

%Before closing we would like to emphasize that Speech Generation enables a whole series of 'textless' applications. Here are a few: discrete speech resynthesis opens up the possibility of \textit{ultralow bitrate codecs}. Here, we achieved bitrates of 150b/sec, which, while 3x higher than text, remains one order of magnitude smaller than the best signal processing-based speech codecs (see \cite{} for a first study using this kinds of units). Conditional speech generation opens up the possibility of text-less \textit{speech to speech translation} which would be useful if the source, target (or both) do not have textual resources. Scaling up this system to larger datasets could enable text-less or audio-augmented end-to-end dialogue-based system. All of these applications would improve naturality  improving AI inclusiveness. 

%This would 
%\textbf{Extension to low resource languages.} All of the tasks we consider here including speech generation can be run in a completely unwritten language, provided enough good quality raw audio is available. This opens up the above mentionned speech and language applications for these languages, improving AI inclusiveness.  

% \appendix
% %%%%%%%%%% Merge with supplemental materials %%%%%%%%%%
% %%%%%%%%%% Prefix a "S" to all equations, figures, tables and reset the counter %%%%%%%%%%

\begin{figure*}[]
     \centering
     \includegraphics[width=0.95\textwidth]{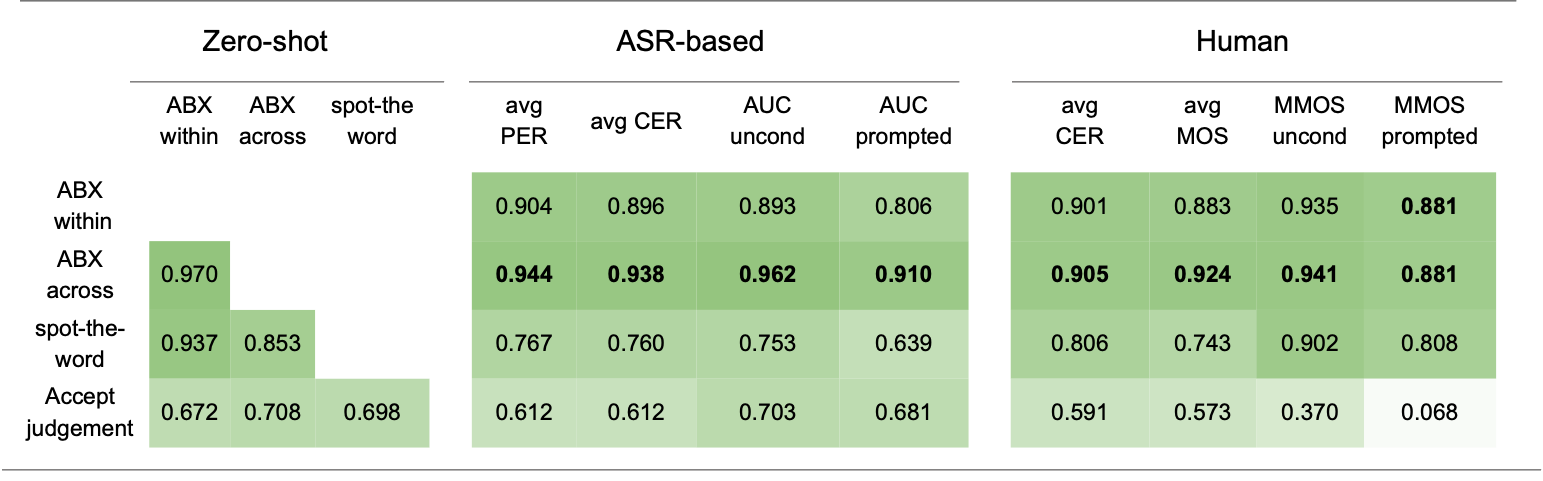}
     \caption{\textbf{Patterns of correlations between the zero-shot metrics and the automatic and human metrics.} Color scale indicates strength of the Pearson correlation coeficient (we used negative MOS and MMOS to enforce less is better for all metrics).}
     \label{fig:zerocorrel}
\vspace{-1em}
 \end{figure*}

\begin{figure}[h!]
     \centering
     \includegraphics[width=0.49\textwidth]{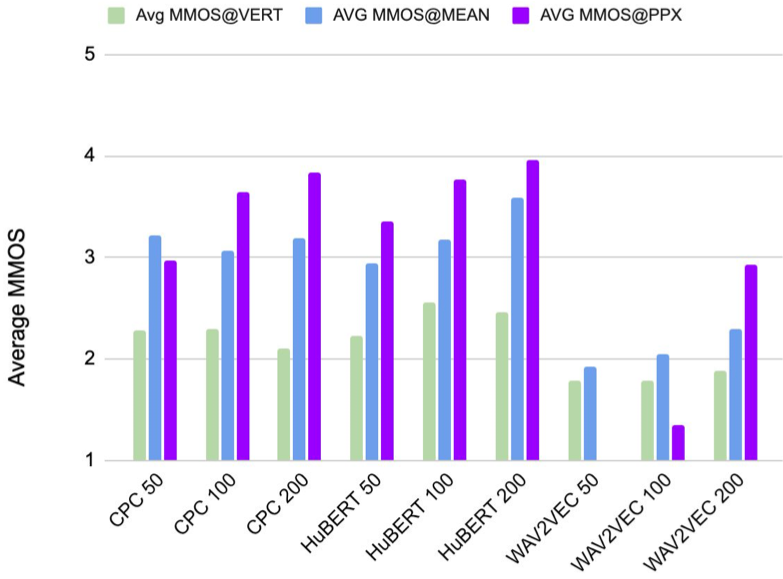}
     \caption{\textbf{MMOS for unconditional (no prompt) and conditional generated speech} sampled at the three reference temperatures (oracle VERT, oracle PPX, and average temperature) (preliminary experiments). }
     \label{fig:MMOSandTEMP}
 \vspace{-1em}
\end{figure}

\vspace{-.2em}
\section{Appendix}\label{sup:sup}
\vspace{-.2em}
\begin{figure}
    \centering
    \includegraphics[width=0.47\textwidth]{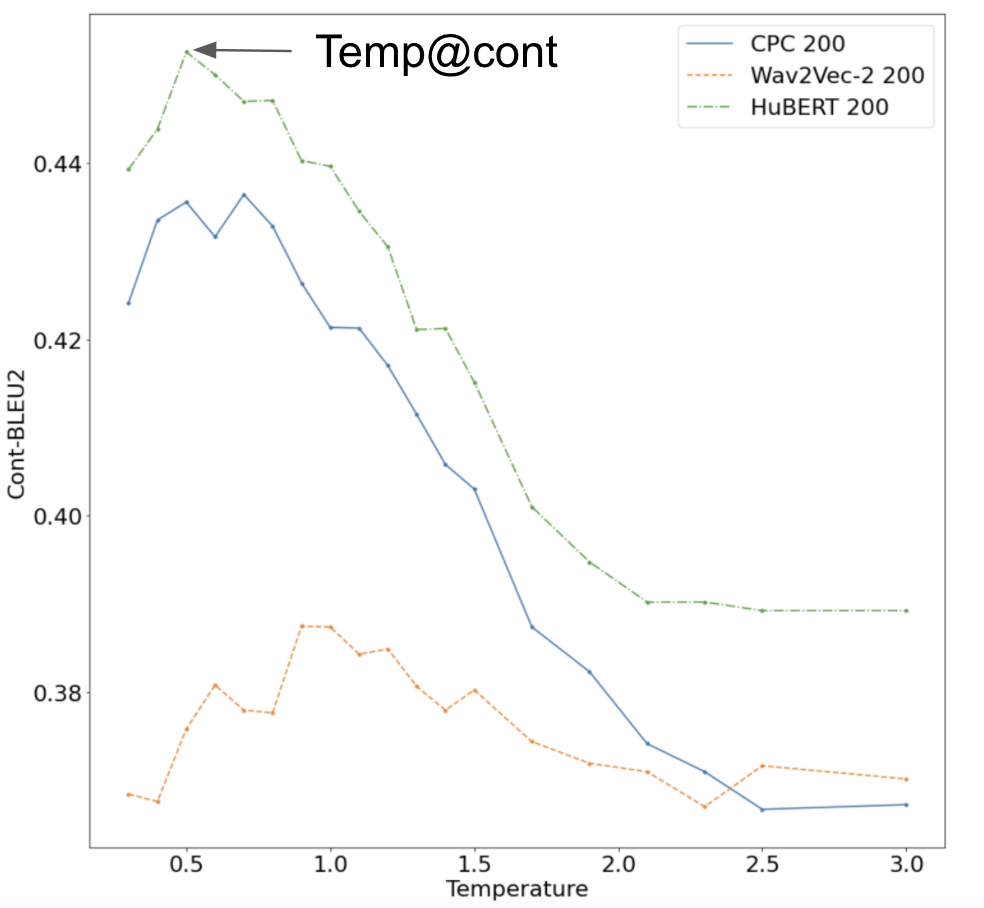}
     \caption{\textbf{Method for selecting the continuation temperature for MMOS judgements}. }
     \label{fig:contTEMP}
 \vspace{-1em}
\end{figure}

\subsection{Zero-shot metrics correlation results}\label{sup:zero}
\vspace{-.3em}

In Figure \ref{fig:zerocorrel}, we present the Pearson correlations between the zero-shot metrics and the human and automatic metrics on downstream tasks. The fact that the ABX metric correlates well with these downstream metrics makes it a useful proxy metric for preliminary model and unit size selection, as it is much less costly than generating TTS output and running human or ASR evaluations.

\subsection{Effect of temperature on outputs}\label{sup:temp}
\vspace{-.3em}

In this section, we describe preliminary experiments we conducted to test the effects of temperature on the generated outputs. As shown in Table \ref{tab:samples}, the temperature defined qualitatively 4 operating zones. With the lowest temperature, we get repetitive outputs, where the system keeps repeating the same few words. At a slightly higher temperature, the system outputs complete sentences, but they are sampled from a narrow set of topics. At the highest temperature, the system utters an unstructured bag of words. In the mid-temperature range, we observe relatively coherent and varied outputs. This is the range we want to select for our systems. As described in Figure 2, the lowest bound was set by using the oracle PPX (temperature range between 0.2 and 0.65. across unsupervised models) and the highest bound by using the oracle VERT (temperature range between 1.1 to 1.4). In Figure \ref{fig:MMOSandTEMP} we present human opinion results for samples from these two temperatures, plus an extra mean temperature falling in between. Humans typically preferred the lower temperature.

In Figure \ref{fig:contTEMP}, we illustrate the continuation method for selecting a single temperature for human meaningfulness judgments in a model-neutral way as explained in Section 3.3. It consists in generating possible continuations of each prompt and computing the BLEU-2 score\footnote{We used NLTK to compute BLEU~\cite{nltk}.} with oracle continuation. The temp@cont temperature is defined as the temperature maximizing this score. Computing these estimates with 10 continuations gave continuation temperatures varying between 0.5 and 0.9 across models and unit sizes. These are the temperatures we used for the MMOS results reported in the main paper.

 %%%%%%%%%%%%%%%%%%%%%%%%%%%%%%%%%%%%
\vspace{-.5em}
\section*{Acknowledgments}
\vspace{-.5em}
We thank Michael Auli and Alexis Conneau for their useful input on wav2vec, and Lior Wolf, Pierre Emmanuel Mazaré and Gargi Gosh for their support for this project. We would also like to thank the reviewers and editors for their thorough review, and constructive feedback.
 %%%%%%%%%%%%%%%%%%%%%%%%%%%%%%%%%%%%

\clearpage

\bibliography{tacl2018}
\bibliographystyle{acl_natbib}

\clearpage
\appendix
%\widetext
\begin{center}
\textbf{\large Supplementary Materials}
\end{center}

%%%%%%%%%% Merge with supplemental materials %%%%%%%%%%
%%%%%%%%%% Prefix a "S" to all equations, figures, tables and reset the counter %%%%%%%%%%
\setcounter{equation}{0}
\setcounter{figure}{0}
\setcounter{table}{0}
\makeatletter
\renewcommand{\theequation}{S\arabic{equation}}
\renewcommand{\thefigure}{S\arabic{figure}}
\renewcommand{\thetable}{S\arabic{table}}
\renewcommand{\bibnumfmt}[1]{[S#1]}
\renewcommand{\citenumfont}[1]{S#1}
\renewcommand{\thesection}{S\arabic{section}}

Here, we provide supplementary information not appearing in the TACL version for lack of space.
\section{Implementation Details}\label{sup:details}

This section provides information about model training.
\subsection{Speech Decoder}\label{sup:pTTS}
We train all speech decoder models referred in Section~\ref{sec:pTTS} on 8 32-GB GPUs using data distributed training with a batch size of 32 per GPU for 500 epochs. We compute validation loss every 5000 steps, and select model with the lowest loss. For chunking (described in Section~\ref{sec:pTTS}), we initialize the chunk size to 50 and increment it by 5 per epoch. Note that we only do chunking of Unit-To-Speech scenario only.

%\subsection{Zero-shot evaluations}\label{sup:zeroeval}
%\subsubsection{ABX}
%\subsubsection{bitrate}
%\subsubsection{spot-the-word}
%\subsubsection{acceptability}

\subsection{Training of ASR Models for evaluation}\label{sup:asreval}
\subsubsection{Frozen ASR Model}\label{sup:asreval_wav2vec960h}
The frozen ASR model is trained using \textsc{large} wav2vec model architecture with Connectionist Temporal Classification (CTC) loss \cite{graves2006ctc} from scratch (not using the pretrained model) on LibriSpeech 960hours dataset.
\subsubsection{Frozen Phoneme Recognition Model}
The frozen phone recognition model is trained using \textsc{base} wav2vec model architecture with Connectionist Temporal Classification (CTC) loss \cite{graves2006ctc} from scratch on LibriSpeech 960hours dataset. We use g2p-en \cite{g2pE2019} for obtaining gold phoneme transcriptions.

\subsubsection{Fitted ASR Model}
For speed of training, we design a \textsc{small} version of wav2vec architecture with a Transformer encoder of 6 layers, 4 attention heads, embedding of size 256 and FFN of size 1024. The model is always trained with CTC loss using synthesized speech from the speech decoder.

\section{Supplementary Results}\label{sup:res}
In Section \ref{sup:fit}, we present the full set of metrics used the resynthetized task, and in section \ref{sup:gene}, those for the generation task.

\subsection{Speech Resynthesis Results on fitted ASR metrics}\label{sup:fit}
In the table \ref{tab:appendix_fitted_asr} we present the full set of metrics that we have used to evaluate synthesized speech using fitted ASR models. All the speech synthesis models are trained on LJSpeech and the synthesized speech is evaluated on LibriSpeech and LJ Speech. Note that we always use dev\_clean set for LibriSpeech, and a random hold-out set of 1000 samples for LJ Speech that were not seen during training or validation.

\begin{table*} [ht!]
    \centering
    \small
    \caption{Full results using fitted ASR and Phoneme Recognition models of three unsupervised unit discovery models as a function of number of quantized units on 4 sets of metrics: PER and CER without LM \& Lexicon (first 2 columns), and WER and CER with LM \& Lexicon (last 2 columns). For comparison we also show the metrics for our supervised topline system.}
    \vspace{0.7em}
    \label{tab:appendix_fitted_asr}
    \begin{tabular}{l|rr|rr|rr|rr}
    \toprule
        \multicolumn{1}{c}{\bf Systems} & \multicolumn{2}{c}{\bf PER} & \multicolumn{2}{c}{\bf CER} & \multicolumn{2}{c}{\bf WER} & \multicolumn{2}{c}{\bf CER (with LM, Lex)}  \\
        \cmidrule(l{3pt}r{3pt}){1-9}
        & LJ & LS & LJ & LS & LJ & LS & LJ & LS \\
        \midrule
        Gold Text + TTS & 8.47 & 11.95 & 5.32 & 10.75 & 20.56 & 17.60 & 9.44 & 8.62 \\
        Pretrained ASR + TTS & - & - & 8.80 & 11.88 & 23.86 & 21.10 & 11.28 & 10.34 \\
        \midrule
        LogMelFbank + KM50 & 34.32 & 60.76 & 25.29 & 48.54 & 69.36 & 92.52 & 39.62 & 63.56 \\
        LogMelFbank + KM100 & 24.94 & 50.58 & 24.01 & 47.40 & 65.27 & 92.13 & 36.43 & 62.64 \\
        LogMelFbank + KM200 & 22.55 & 53.21 & 20.99 & 46.37 & 57.61 & 90.98 & 30.72 & 59.01 \\
        \midrule
        CPC + KM50  & 16.04 & 26.72 & 13.27 & 27.61 & 39.16 & 63.10 & 18.52 & 33.26 \\
        CPC + KM100 & 19.23 & 32.05 & 12.48 & 25.93 & 35.77 & 58.09 & 16.63 & 29.57 \\
        CPC + KM200 & 15.92 & 26.78 & \bf9.98 & 23.22 & \bf31.23 & 53.56 & \bf14.30 & 26.73 \\
        \midrule
        HuBERT L6 + KM50  & 15.37 & 24.58 & 12.72 & 25.66 & 39.15 & 58.31 & 18.30 & 28.91 \\
        HuBERT L6 + KM100 & 16.57 & 25.83 & 11.74 & 24.56 & 31.27 & 51.67 & 14.60 & 26.36 \\
        HuBERT L6 + KM200 & 15.88 & 24.11 & 11.20 & \bf21.62 & 32.91 & \bf50.53 & 16.00 & \bf26.20 \\
        \midrule
        wav2vec L14 + KM50  & 23.89 & 36.18 & 27.60 & 39.55 & 71.19 & 86.27 & 44.52 & 57.51 \\
        wav2vec L14 + KM100 & 16.62 & 27.76 & 17.11 & 30.67 & 48.84 & 71.53 & 24.51 & 38.28 \\
        wav2vec L14 + KM200 & \bf13.79 & \bf24.06 & 11.25 & 23.54 & 34.55 & 55.53 & 15.96 & 27.48 \\
        \bottomrule
    \end{tabular}
\end{table*}

\subsection{Generation Task}\label{sup:gene}

Table \ref{tab:full_mmos} shows the results of the meaninfulness opinion score (MMOS) detailing the three temperature settings (oracle PPX, oracle VERT and the average temperature between these two endpoints). As seen on Figure \ref{fig:MMOSandTEMP}, human judgements are usually higher for the oracle PPX than the other two temperatures, and the pattern of results across systems is globally comparable across temperatures. Figure \ref{fig:MMOScorrel} shows the table of correlation coefficients across human and automatic metrics.

\begin{figure*}[h]
    \centering
    \includegraphics[width=0.99\textwidth]{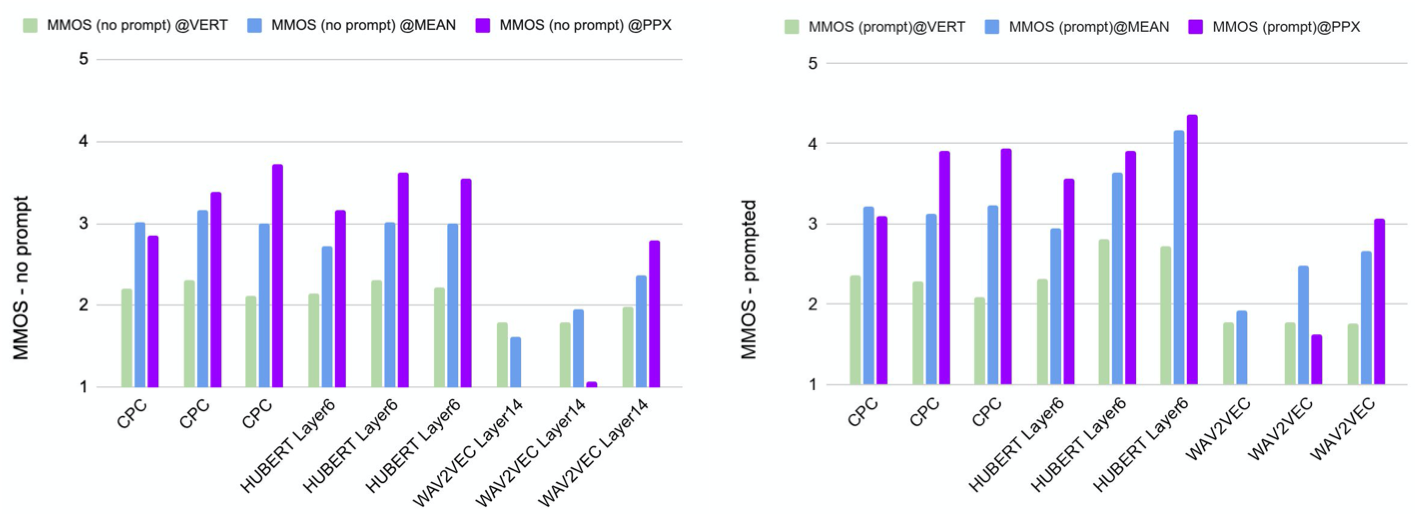}
    \caption{\textbf{MMOS for unconditional (no prompt) and conditional generated speech} sampled at the three reference temperatures (oracle VERT, oracle PPX, and average temperature). }
    \label{fig:MMOSandTEMP}
\end{figure*}

\begin{figure*}[h]
    \centering
    \includegraphics[width=0.99\textwidth]{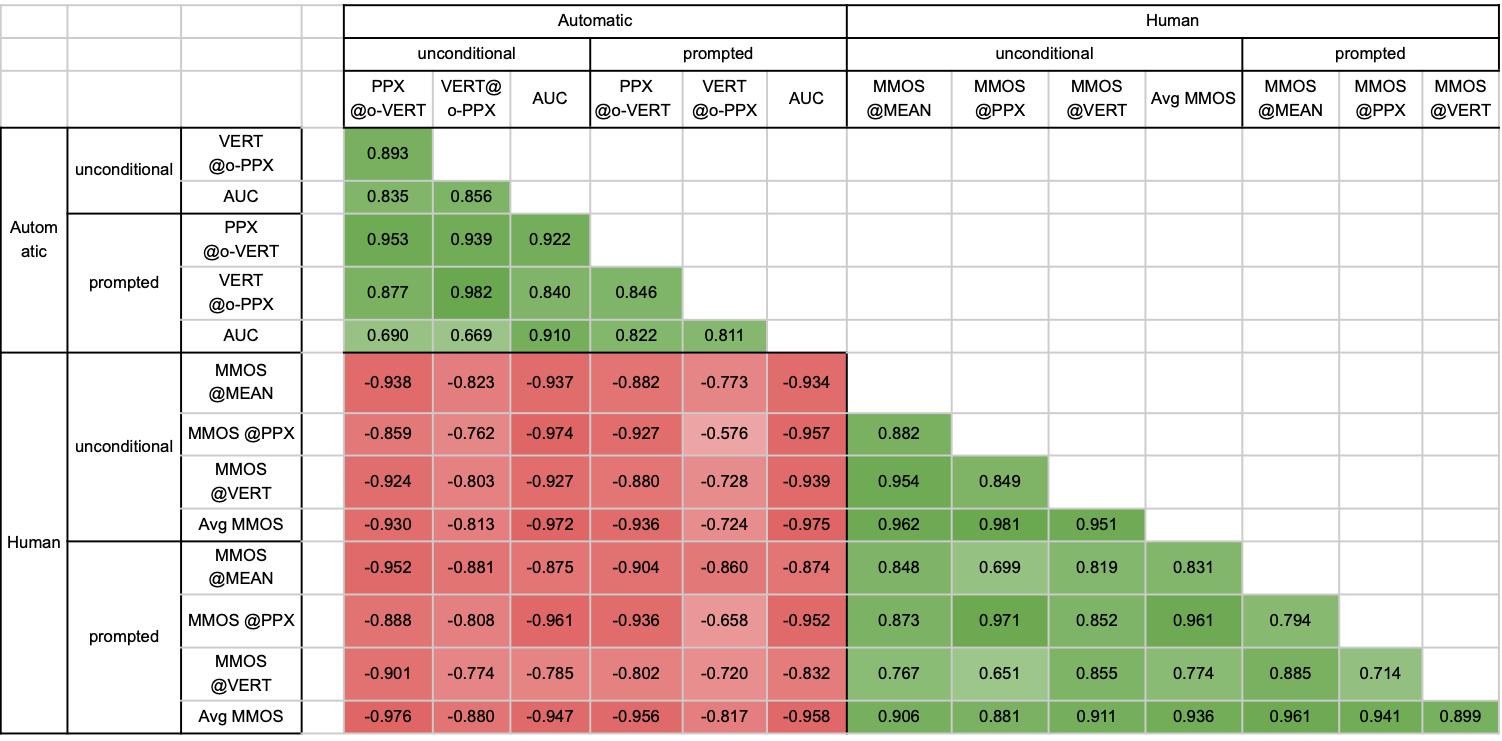}
    \caption{\textbf{Patterns of correlations between automatic and human metrics for generation.} Color scale indicates strength and direction of the Spearman correlation coeficient. }
    \label{fig:MMOScorrel}
\end{figure*}

\begin{table*}[t]
\centering
\caption{Full Human evaluation MMOS results for three unsupervised models and 3 unit sizes. For each model and unit size, samples were generated using three different temperatures corresponding to: VERT, PPX and the average temperature of both.}
\label{tab:full_mmos}
\begin{tabular}{l c || c c  c | c  c c}
\hline
\mc{2}{c||}{Systems}  & \mc{6}{c}{MMOS}       \\
\hline
Encoder     & Nb     &\mc{3}{c|}{\ux{unconditional}}       &\mc{3}{c}{\ux{prompt}}     \\
architect.  & units  & PPX     & VERT    &    MEAN   & PPX      & VERT   &     MEAN \\
\hline
\mc{2}{l@{\hs{0.1}}||}{\textit{Controls}}&&&&&      \\
\mc{2}{l@{\hs{0.1}}||}{oracle text}& - &   -  &  -  & - & - & 4.44 \\
\mc{2}{l@{\hs{0.1}}||}{ASR + LM}& 3.72 & 3.23 &	3.57 & 3.24 & 3.02 & 3.24 \\
\hline
\mc{2}{l@{\hs{0.1}}||}{\textit{Unsupervised}}&&&&& \\
CPC         & 50  & 2.85 & 2.20 & 3.02 & 3.10 & 2.35 & 3.22 \\
CPC         & 100 & 3.38 & 2.31 & 3.16 & 3.90 & 2.28 & 3.12 \\
CPC         & 200 & 3.72 & 2.11 & 3.00 & 3.94 & 2.08 & 3.23 \\
HuBERT-L6   & 50  & 3.16 & 2.15 & 2.72 & 3.56 & 2.31 & 2.94 \\
HuBERT-L6   & 100 & 3.62 & 2.31 & 3.02 & 3.90 & 2.81 & 3.64 \\
HuBERT-L6   & 200 & 3.55 & 2.22 & 3.00 & 4.36 & 2.71 & 4.16 \\
Wav2vec-L14 & 50  & -    & 1.79 & 1.62 & -    & 1.77 & 1.92 \\
Wav2vec-L14 & 100 & 1.07 & 1.79 & 1.95 & 1.62 & 1.77 & 2.48 \\
Wav2vec-L14 & 200 & 2.80 & 1.99 & 2.37 & 3.06 & 1.76 & 2.65 \\
\hline
\end{tabular}
\end{table*}
\end{document}